\documentclass[authoryear, preprint,1p,times]{elsarticle}
\usepackage[table,xcdraw]{xcolor}
\usepackage{graphicx}
\usepackage{svg}
\usepackage[separate-uncertainty = true,multi-part-units=single]{siunitx}
\usepackage{subcaption}
\usepackage{caption}
\usepackage{booktabs}
\usepackage{amsmath}
\usepackage{multirow}
\usepackage{bm}
\usepackage{placeins}
\usepackage{amssymb}  
\usepackage{wasysym}   

\journal{}

\renewcommand{\mathrm}[1]{#1}
\newcommand{\reals}{{\mbox{\bf R}}}

\newcommand{\oraclesymb}{\textcolor{black}{$\blacktriangle$}}
\newcommand{\constsymb}{\textcolor{green}{$\blacksquare$}}
\newcommand{\cubesymb}{\textcolor{blue}{$\blacklozenge$}}
\newcommand{\zerosymb}{\textcolor{red}{$\CIRCLE$}}

\renewcommand{\quote}[1]{``#1''}

\begin{document}

\begin{frontmatter}

\title{Predicting center of mass position in non-cyclic activities: The influence of acceleration, prediction horizon, and ground reaction forces}

\author[1]{Mohsen Alizadeh Noghani\corref{cor1}}

\author[1]{Edgar Bolívar-Nieto}

\affiliation[1]{organization={Aerospace and Mechanical Engineering Department, University of Notre Dame},
            city={Notre Dame},
            postcode={46556}, 
            state={IN},
            country={U.S.}}

\cortext[cor1]{Corresponding author. Email: malizade@nd.edu.}

\begin{abstract}

 The whole-body center of mass (CoM) plays an important role in quantifying human movement. Prediction of future CoM trajectory, modeled as a point mass under influence of external forces, can be a surrogate for inferring intent. Given the current CoM position and velocity, predicting the future CoM position by forward integration requires a forecast of CoM accelerations during the prediction horizon. However, it is unclear how assumptions about the acceleration, prediction horizon length, and information from ground reaction forces (GRFs), which provide the instantaneous acceleration, affect the prediction.
We study these factors by analyzing data of 10 healthy young adults performing 14 non-cyclic activities. We assume that the acceleration during a horizon will be 1) zero, 2) remain constant, or 3) converge to zero as a cubic trajectory, and  perform predictions for horizons of 125 to 625 milliseconds. We quantify the prediction performance by comparing the  position error and accuracy of identifying the main direction of displacement against trajectories obtained from a whole-body marker set. For all the assumed accelerations profiles, position errors grow quadratically with horizon length ($R^2 > 0.930$) while the accuracy of the predicted direction decreases linearly ($R^2>0.615$). 
 Post-hoc tests reveal that the constant and cubic profiles, which utilize the GRFs, outperform the zero-acceleration assumption in position error ($p<0.001$, Cohen's $d>3.23$) and accuracy  ($p<0.034$, Cohen's $d>1.44)$ at horizons of 125 and 250\,$\SI{}{\milli\second}$. The results provide evidence for benefits of incorporating GRFs into predictions and point to 250\,$\SI{}{\milli\second}$ as a threshold for horizon length in predictive applications.

\end{abstract}

\end{frontmatter}

\section{Introduction}

The whole-body center of mass (CoM) is a fundamental marker of human movement, with applications including quantifying balance and predicting foot placement \citep{bruijn2018control}. Previous research has determined conditions for forward fall, backward fall, and recovery while standing based on the CoM state (i.e., its position and velocity) \citep{pai1997center, iqbal2000predicted}. Models have been proposed to predict the changes in the width of the next step as a function of the current CoM state due to the natural gait variability \citep{wang2014stepping},  and due to perturbations in mediolateral and anteroposterior directions \citep{vlutters2016center} . The extrapolated center of mass (XCoM)  determines the margin of stability, defined as the distance between the XCoM and the base of support \citep{hof2005condition,hof2008extrapolated}. The resulting spatial and temporal metrics \citep{bruijn2013assessing, buurke2023comparison} have been used to study stability during scenarios such as healthy walkers navigating cluttered terrain \citep{moraes2007validating}, above-knee amputees walking on a treadmill \citep{hof2007control}, and transtibial amputees traversing rough surfaces \citep{curtze2011rough}. 

CoM kinematics can also inform stabilizing control action of wearable robots, such as adjusting the hip angle of a lower-limb exoskeleton to return the XCoM to a nominal state in response to perturbations \citep{zhang2018design}. Other researchers have designed controllers for a prosthetic ankle that stabilize the user's gait by feedback control of the CoM velocity \citep{kim2015once} and acceleration \citep{kim2017step} in the mediolateral direction. For ankle exoskeletons, CoM velocity feedback can produce stabilizing torque after anteroposterior perturbations \citep{bayon2022cooperative, afschrift2023assisting}. Interestingly, CoM velocity feedback results in faster stabilization compared to feedback from muscle activity when providing ankle torque, as the changes in the former are more immediate \citep{beck2023exoskeletons}.

 Despite the applications of the instantaneous CoM state in human biomechanics, research on prediction of future CoM position trajectory has focused on planning and control of legged robots, particularly humanoids, often in tandem with determining foot placements to ensure stability of the robot \citep{kuindersma2016optimization, wensing2024optimization}. For humans, predictions of whole-body CoM position can be seen as proxies for their intended direction of movement, a key factor in wearable robotics, such as powered prosthetic legs \citep{azocar2020design}. For example, during walking, the main component of CoM motion is forward translation with relatively small vertical and lateral oscillations \citep{tesio2019motion} while during transition from sitting to standing, the CoM experiences horizontal and vertical displacements \citep{van2021compensation}.  Despite this, research in prediction of the CoM position has been limited. A recent work has applied a temporal convolutional neural network (TCN) to body segment angles to predict the CoM position and velocity during walking on a treadmill after perturbations \citep{leestma2024data}. However, 
 utilizing ground reaction forces (GRFs), which are partially available in some lower-limb wearable robots \citep{azocar2020design, divekar2024versatile}, and provide information about the CoM acceleration, has not been explored.

 To enable applications that incorporate prediction of CoM position, it is crucial to establish how the prediction error is affected by different horizon lengths, assumptions about CoM dynamics, and as a result of incorporating information from GRFs. To address these gaps in the literature, in this work, we predicted the CoM position of subjects performing a set of 14 non-cyclic activities. We defined three profiles for the acceleration of the CoM in the future and compared their performance using four prediction error metrics across five horizon lengths. We hypothesized that (1) incorporating GRFs would improve prediction of CoM position compared to assuming zero acceleration, and (2) the prediction would become less accurate and more uncertain in the longer time horizons. 

\section{Methods}

\subsection{Data collection and processing}

Data of ten healthy participants ($n=10$, mean~±~SD: age 22.9 $\pm$ 3.5 years, height \SI{1.73(0.083)}{\meter}, mass  \SI{63.66(9.79)}{kg}), including five males and five females, with no self-reported pathologies affecting balance or gait, was collected. 
The study was approved by the University of Notre Dame Institutional Review Board (Protocol ID: 24-01-8277) and the participants gave written informed consent.

\begin{figure}
  \centering
  \includegraphics[width=0.9\linewidth]{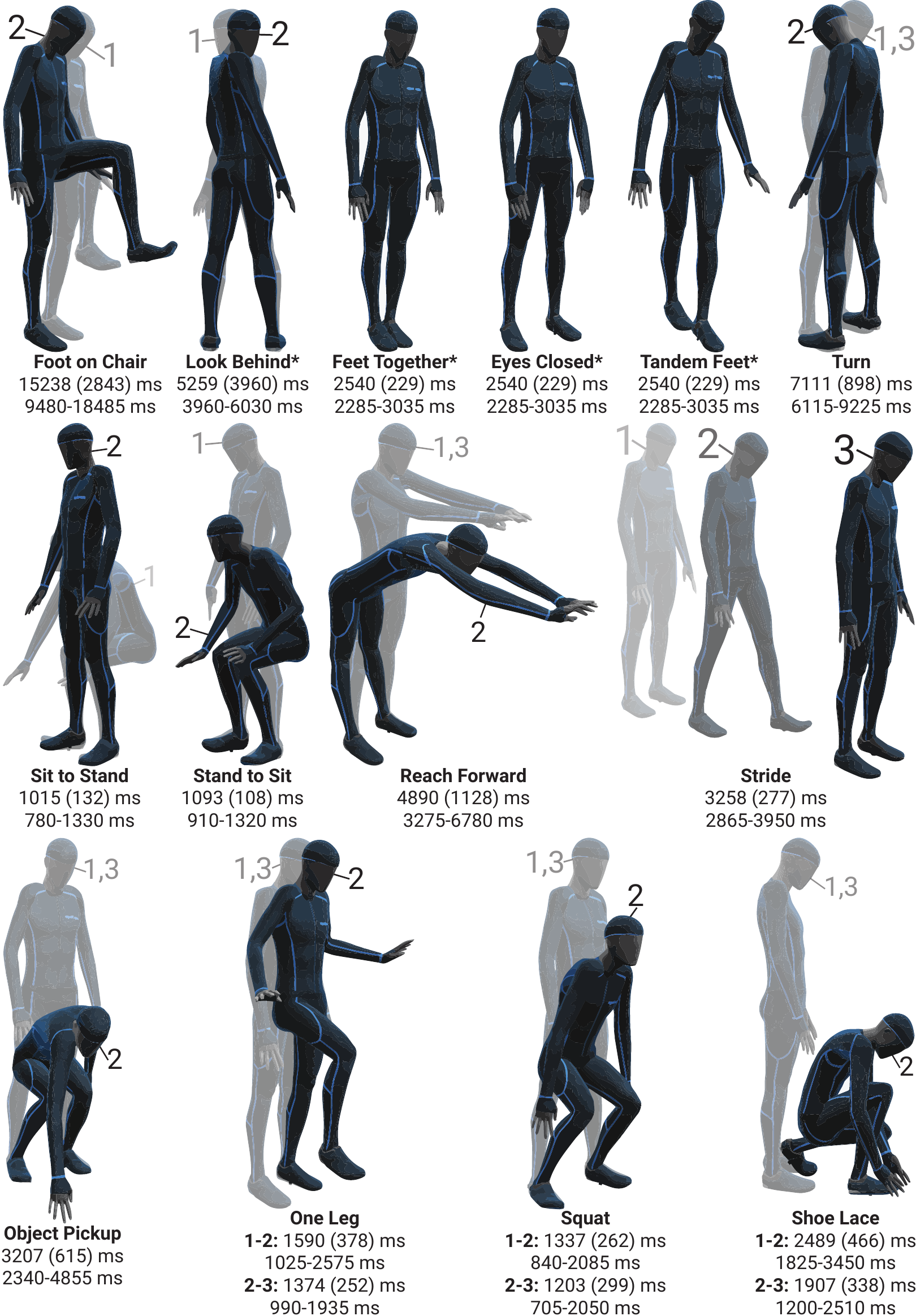} 
  \caption{The activities and statistics of their durations, reported in milliseconds as mean (standard deviation) and minimum-maximum. The numbers 1,2,3 correspond to consecutive snapshots of the movements. One Leg, Squat, and Shoe Lace were split into start (from 1 to 2) and return (from 2 to 3) phases. The activities in the Static group are marked with (*). A written description of each activity is available in \ref{sect:activities}.}
  \label{fig:activities}
\end{figure}

Fifty-seven markers were placed on the participants according to the Biomech (57) whole-body markerset \citep{optitrack_biomech57}. The participants performed 14 activities (Fig.~\ref{fig:activities}, \ref{sect:activities}), adapted from the Berg Balance Scale test \citep{berg1992measuring}, each repeated three times, over a walkway equipped with force plates. To enable natural motion, the activities were performed at the participant's own pace.
Four activities (Look Behind, Feet Together, Eyes Closed, Tandem Feet) were qualitatively classified as \quote{Static}, given the small displacements of the body segments. 
The force plates (Kistler Instrumente AG, Winterthur, Switzerland) collected 3D GRFs at \SI{1000}{\hertz} while a twelve-camera motion capture system (PrimeX22, NaturalPoint Inc., OR, USA) recorded the marker positions at \SI{200}{\hertz}. Gap-filling and zero-lag low-pass filtering (\SI{6}{\hertz}) of the marker data were performed in Motive 3.0 (NaturalPoint Inc., OR, USA).

Inverse kinematics analysis and processing of force plate data were performed in MATLAB (MathWorks Inc., MA, USA) and OpenSim \citep{delp2007opensim}. 
To determine the foot and force plate contact events, we manually labeled the points where each force plate's vertical force's began to rise or leveled off. 
The GRF data in the absence of contact was clamped to zero to mitigate the effect of noise and crosstalk, and subsequently  lowpass-filtered at \SI{20}{\hertz} (5th order Butterworth filter) and downsampled to \SI{200}{\hertz} to match the sample rate of marker data. To emphasize segments of data that included transient CoM dynamics, data of One-Leg, Squat, and Shoe-Lace activities were split into two phases: start (e.g., lifting the leg) and return (e.g., lowering the leg) (Fig.~\ref{fig:activities}). These were analyzed separately, and the data between them (e.g., standing on one leg for 10 seconds) were excluded. 
We adjusted a full-body OpenSim model \citep{rajagopal2016full} by adding a head-and-neck segment, connected to the torso with a ball-and-socket joint, with its mass and moment of inertia calculated according to anthropometric tables \citep{dumas2007adjustments}. 
The OpenSim model was scaled according to the subject's mass and anthropometric measurements based on the markers. Using the inverse kinematics results, the whole-body CoM position and velocity were computed by OpenSim's BodyKinematics analysis tool \citep{seth2011opensim,sherman2011simbody}.

\subsection{CoM prediction}

The sum of all external forces acting on a body are equal to its mass and the second derivative of its CoM position. Thus, we modeled the CoM dynamics as a double-integrator system with the following discrete-time state space representation,
\begin{gather}
    x[k] = A x[k-1] + B u[k-1],\label{eq:disc_state}
\end{gather}
where the state vector $x \in \reals^6$ contains the 3D components of the CoM position and velocity, and the input $u \in \reals^3$ is the acceleration of the CoM computed using GRFs measured by the force plates. $A$ is the state transition matrix, and $B$ represents the influence of control inputs on the states, both obtained from discretizing the continuous dynamics at a sample time of $\Delta T = \SI{5}{\milli\second}$ with the zero-order hold assumption (\ref{sect:dynamics}).

\begin{figure}[t!]
    \centering
    \begin{subfigure}{0.4\textwidth}
        \centering
        \includegraphics[width=\linewidth]{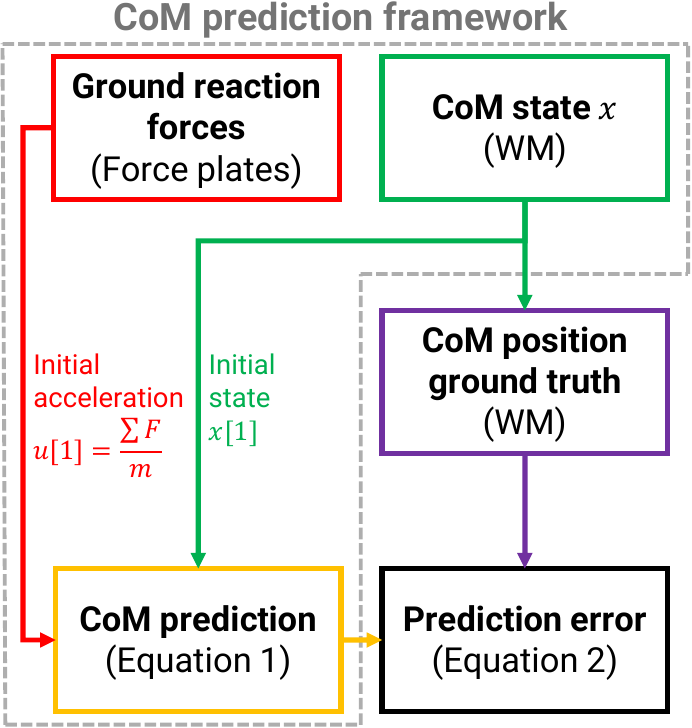}
        \caption{}
        \label{fig:pred_framework}
    \end{subfigure}\hfill
    \begin{subfigure}{0.55\textwidth}
        \centering
        \includegraphics[width=\linewidth]{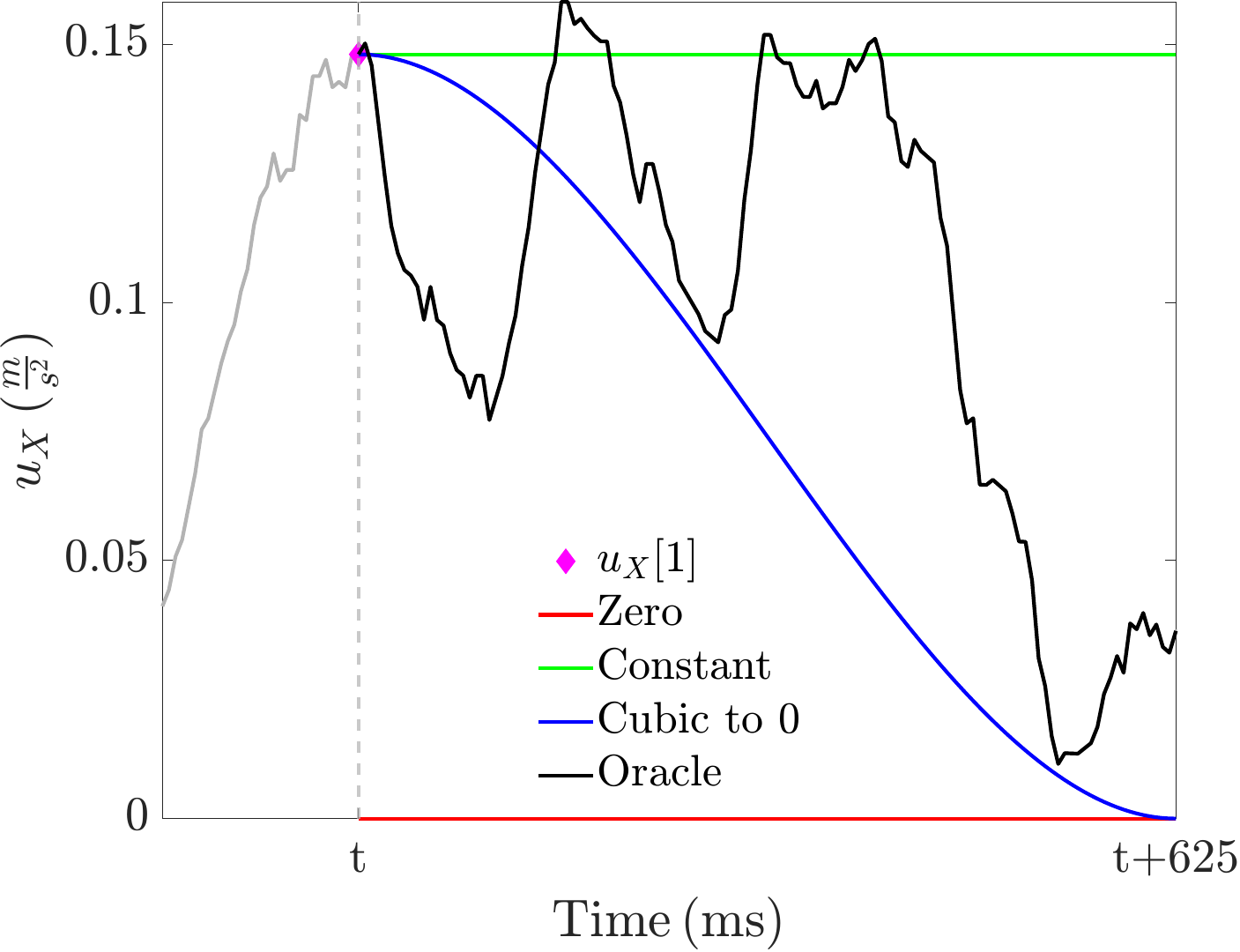}
        \caption{}
        \label{fig:accel_profiles}
    \end{subfigure}
    \caption{(a) The framework for prediction of CoM position in a horizon, which works in two steps: 1) estimating the current CoM state to supply $x[1]$, and 2) predicting the CoM position over the time horizon by propagating the CoM dynamics. WM refers to the estimate from the whole-body marker set computed by OpenSim.  (b) An example of the acceleration profiles for a horizon with the length \SI{625}{\milli\second}. The vertical dashed line shows the start of the horizon.}
\end{figure}

To predict the CoM position trajectory over a time horizon with length $T$ milliseconds, consisting of $N_s$ samples (Fig.~\ref{fig:pred_framework}), in addition to the initial state, we need a prediction of the acceleration over the horizon, denoted by $u_h[k]\ (k = 1,\hdots,N_s)$. Using the instantaneous acceleration at the first sample of the horizon, $u[1]$, we defined the following acceleration profiles for each of the three axes of motion (Fig.~\ref{fig:accel_profiles}):
\begin{enumerate}
    \item \textbf{Zero}: the acceleration was 0 over the horizon ($ u_h[k] = 0$). This provided a baseline for comparison with the profiles that utilized GRFs for prediction.  
    \item \textbf{Constant (Const)}: the acceleration remained constant over the horizon ($u_h[k] = u[1]$). This is a common assumption in forecasting of time series modeled in state-space \citep{pml2Book}. 
    \item \textbf{Cubic to 0 (Cubic)}: the acceleration started from $u[1]$ and reached 0 at the end of the horizon, following a cubic equation. Motivated by the application of minimum-jerk planning in humanoid robots \citep{van2017real}, the jerk at the start and end of the horizon was 0.
\end{enumerate}
Additionally, the acceleration \textbf{Oracle} was calculated by dividing the future net external forces by mass. We hypothesized that this ideal profile would be a reference for the CoM prediction error achievable by integration of the dynamics.  
The CoM state was predicted by propagating the state of the system using Eq.~(\ref{eq:disc_state}) for the $N_s$ samples in the horizon. Each horizon began at sample $i$ of the recorded data (starting from $i=1$) and included samples up to $i+N_s-1$, and prediction was performed until the end of a horizon coincided with the last sample of the data.

We investigated 5 horizon lengths of $T \in \{125, 250, 375, 500, 625\}$ milliseconds,  corresponding to  $N_s \in \{26, 51, 76, 101, 126 \}$ samples. The minimum length of \SI{125}{\milli\second} was determined by previous results showing that the error was not affected by the choice of profile in shorter horizons \citep{noghani2024prediction}. The time difference between the horizon lengths (\SI{125}{\milli\second}) was selected to be close to muscle reaction times when maintaining standing balance \citep{horak1986central, beck2023exoskeletons}. The maximum length of \SI{625}{\milli\second} was limited by the duration of the shortest recorded data, at \SI{705}{\milli\second}, corresponding to the return phase of Squat (Fig.~\ref{fig:activities}). 

We calculated the prediction error at each sample as the Euclidean distance between the predicted CoM position $\hat{p} \in \reals^3$ and the estimate given by whole-body marker set, $p  \in \reals^3$, 
\begin{gather}
e[k] = \| \hat{p}[k] - p[k] \|_2, \quad k = 1,\ldots,N_s,\label{eq:e_k}
\end{gather}
where $\left\|.\right\|_2$ denotes the $L_2$-norm. To characterize the prediction error, we defined the following metrics:

\begin{enumerate}
    \item \textbf{Average error ($\bm{AE}$)}: the prediction error, $e[k]$, was calculated at each sample $k$ in the horizon. Then, it was averaged across the samples $s$ in the horizon (by dividing by $N_s$), horizons $h$ in each repeat of the activity, repeats $r$ of the activity, and the activities $a$,
    \begin{gather}
        AE = \text{E}_a \text{E}_r \text{E}_h \text{E}_s e[k],\label{eq:AE}
    \end{gather}
    where $\text{E}$ represents taking the sample mean.
    \item \textbf{Maximum error ($\bm{ME}$)}: instead of the average, the maximum was calculated to provide a measure of the worst-case prediction error,
    \begin{gather}
        ME = \max\limits_{a} \max\limits_{r} \max\limits_{h} \max\limits_{s} e[k].\label{eq:ME}
    \end{gather}
    \item \textbf{Average displacement direction accuracy ($\bm{ADA}$)}: in each horizon, we identified the axis of movement (i.e., either $X$, $Y$, or $Z$) with the largest displacement according to the whole-body marker set, denoted by subscript $M$ in Eq.~(\ref{eq:accuracy}). If the predicted displacement was in the same direction as the whole-body marker set, we assigned an accuracy score ($S$) of 1 and 0 otherwise. The results were then averaged.
        \begin{gather}
S =
\begin{cases}
1 & \text{if }\operatorname{sgn}(\hat{p}_M[N_s] - p_M[1]) = \operatorname{sgn}({p}_M[N_s] - p_M[1]), \\
0 & \text{otherwise},
\end{cases}\label{eq:accuracy}\\
        ADA = \text{E}_a \text{E}_r \text{E}_h S.
    \end{gather}

This evaluated the accuracy of prediction in identifying the direction the participant intended to move. We note that the Static activities were excluded due to the small CoM displacement.

\item \textbf{Minimum displacement direction accuracy ($\bm{MDA}$)}: instead of the average, the minimum was taken over the repeats and activities. This provided a worst-case metric of displacement direction accuracy for each subject,
\begin{gather}
   MDA = \min\limits_{a} \min\limits_{r} \text{E}_h S.
\end{gather}
    
\end{enumerate}

\subsection{Statistical analysis}

Performing the analysis described in the previous subsection resulted in $N=10$ samples for each combination of prediction error metric, horizon length, and acceleration profile. Q-Q plots were used to evaluate the normality of the data.
Using F-tests, we first compared cubic and quadratic equations describing the relationship between the four error metrics and horizon length $T$,
with the null hypothesis that there is no cubic trend \citep{maxwell2017designing}. Because the null was not rejected for any of the tests ($p > 0.235$), we subsequently compared quadratic and linear fits. 95\% confidence intervals (CI) were computed to quantify the uncertainty of the estimates. Due to non-equal variance at different horizon lengths, the parameters of the linear and quadratic models were estimated by weighted least squares.

To compare the acceleration profiles, one-way Welch's analysis of variance (ANOVA) was performed at each horizon length, followed by Welch's t-tests with the Bonferroni correction in case of significant main effects.
Significance level was set to $\alpha=0.05$, and Cohen's d ($d$) was used to quantify effect sizes of the pairwise tests \citep{liu2021t}. Thresholds for small, medium, and large effects sizes were 0.2, 0.5, and 0.8, respectively \citep{cohen2013statistical}. All statistical analyses were performed in R.

\section{Results}

\begin{figure}[tb!]
    \centering
    \begin{subfigure}{0.49\textwidth}
        \centering
        \includegraphics[width=\linewidth]{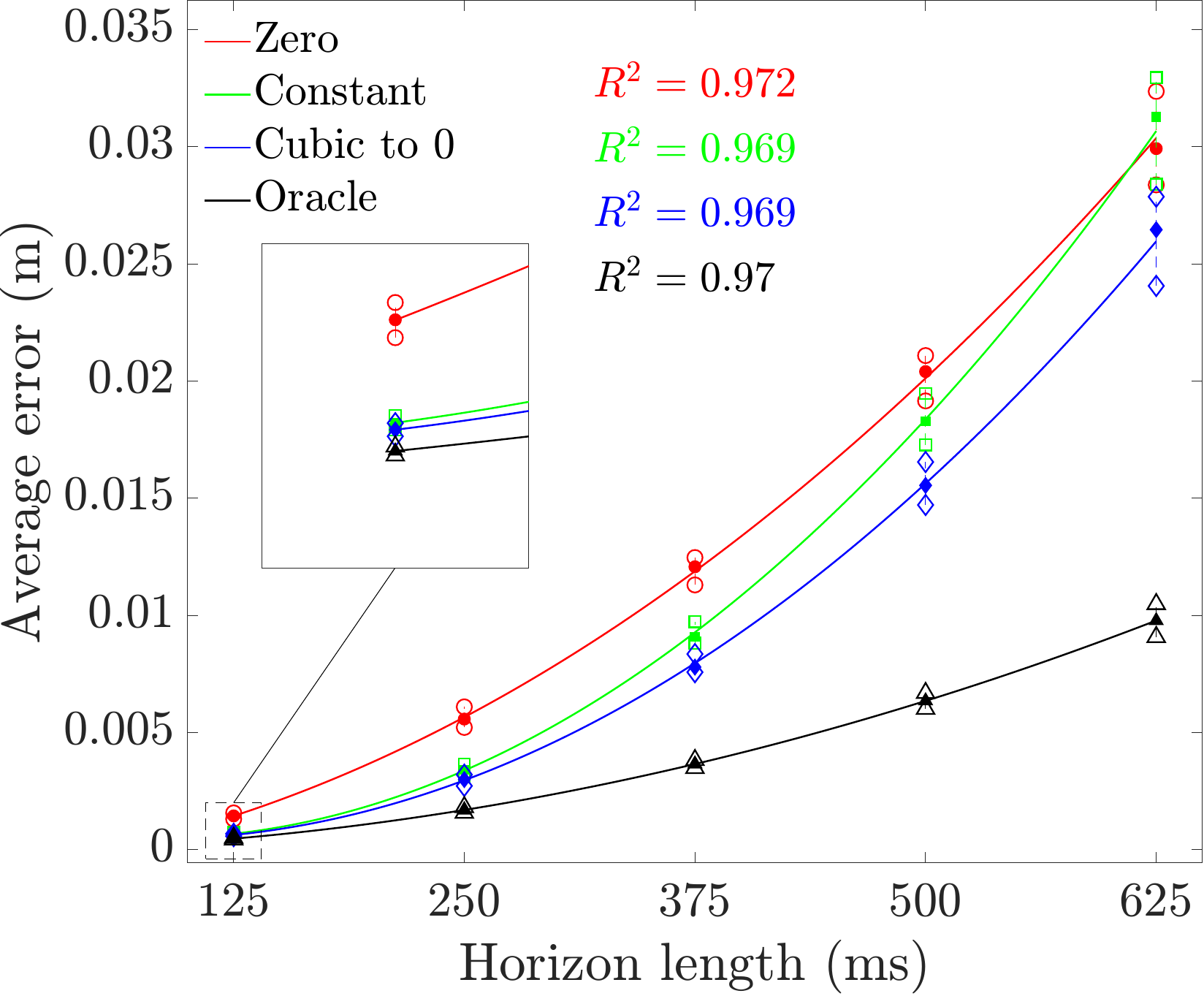}
        \caption{}
        \label{fig:AE_fit}
    \end{subfigure}\hspace{0.2cm}
    \begin{subfigure}{0.475\textwidth}
        \centering
        \includegraphics[width=\linewidth]{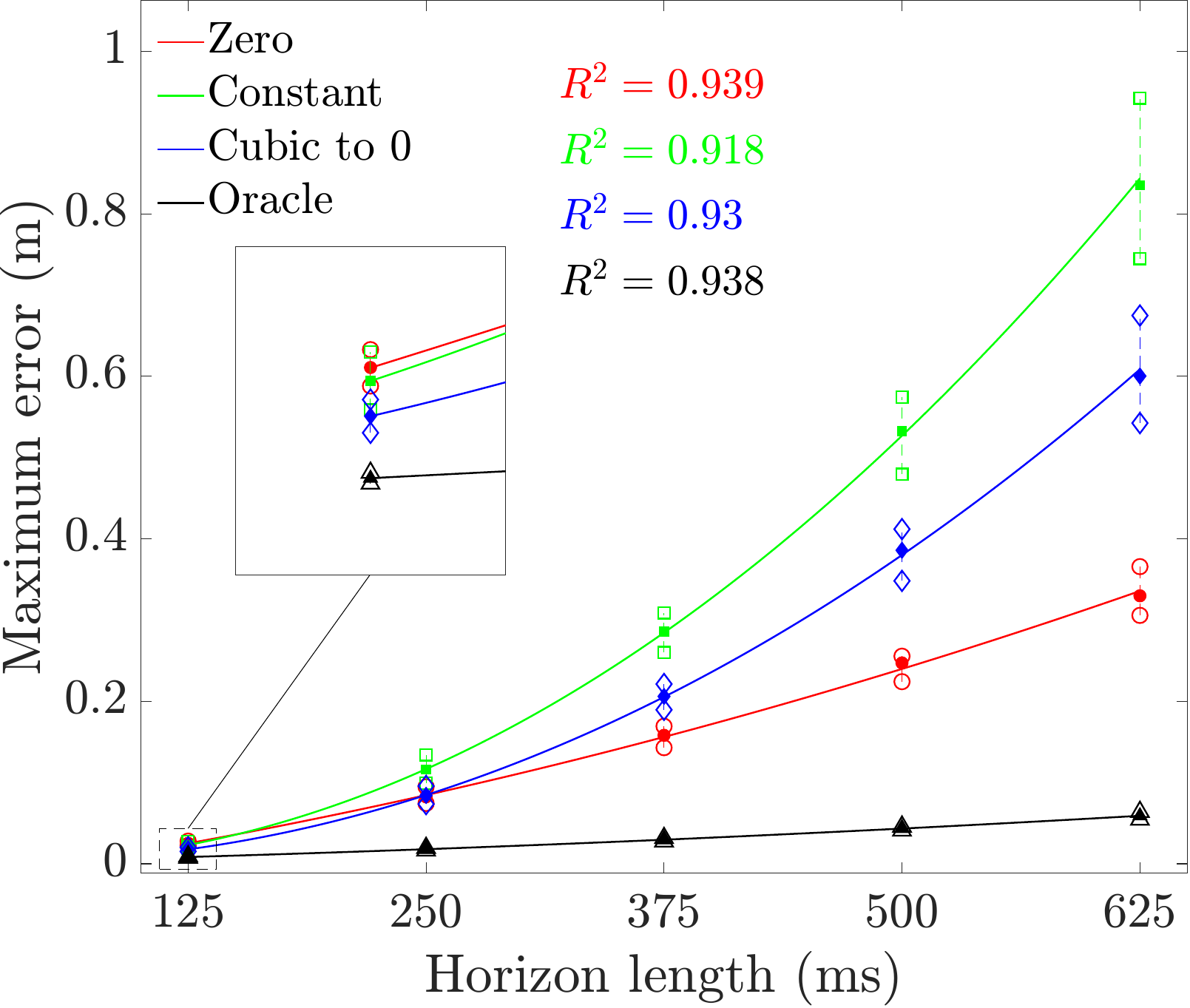}
        \caption{}
        \label{fig:ME_fit}
    \end{subfigure}
    \caption{The fitted curves for (a) the average error ($AE$), and (b) the maximum error ($ME$). The solid markers (\zerosymb\ for Zero, \constsymb\ for Constant, \cubesymb\ for Cubic to 0, and \oraclesymb\ for Oracle) show the mean of the data points, while the hollow markers represent the 95\% CIs.}
\end{figure}

The F-tests revealed a quadratic trend for $AE$ in all acceleration profiles ($p<0.001$), with the $R^2$ values of close to 0.97 for all models (Fig.~\ref{fig:AE_fit}). ANOVA identified that the acceleration profile had a significant effect on $AE$ at all horizons ($p<0.001$); while pairwise tests detected differences with large effect sizes between Const-Zero and Zero-Cubic in shorter horizons, the null hypothesis was not rejected for Const-Cubic at any $T$ (Table~\ref{tab:AE_ME_pairs}). The comparisons between Oracle and the three other profiles were significant at all horizon lengths ($p<0.001$, $d>2.35$).

Similar to $AE$, the F-tests for $ME$ indicated that the quadratic models provided a significantly better fit than linear ($p<0.014$), with $R^2 > 0.918$ (Fig.~\ref{fig:ME_fit}). The results of ANOVA showed that the choice of acceleration profile influenced $ME$ for all $T$, and post-hoc tests identified differences between Zero-Const and Zero-Cubic in three of the horizon lengths; however, there were no significant differences between Const-Cubic (Table~\ref{tab:AE_ME_pairs}). Meanwhile, $ME$ of Oracle was significantly lower than the other profiles at all horizon lengths ($p<0.001$, $d>3.12$).

For $ADA$, the quadratic models did not perform better than the linear models ($p>0.131$ in the F-tests), with the latter achieving $R^2$ values ranging from 0.671 for Oracle to 0.902 for Constant (Fig.~\ref{fig:ME_fit}). ANOVA indicated differences between the four profiles for all levels of $T$ ($p<0.001$), and the post-hoc tests found differences between Zero-Const at $T=\SI{125}{\milli\second}$ and between Zero-Cubic at all values of $T$; no significant difference was found between Const and Cubic profiles (Table~\ref{tab:ADA_MDA_pairs}). With the exception of the Oracle-Cubic pair at \SI{125}{\milli\second} ($p=0.144$), other comparisons with the Oracle were significant ($p<0.039$, $d>1.38$).

\begin{figure}[b!]
    \centering
    \begin{subfigure}{0.49\textwidth}
        \centering
        \includegraphics[width=\linewidth]{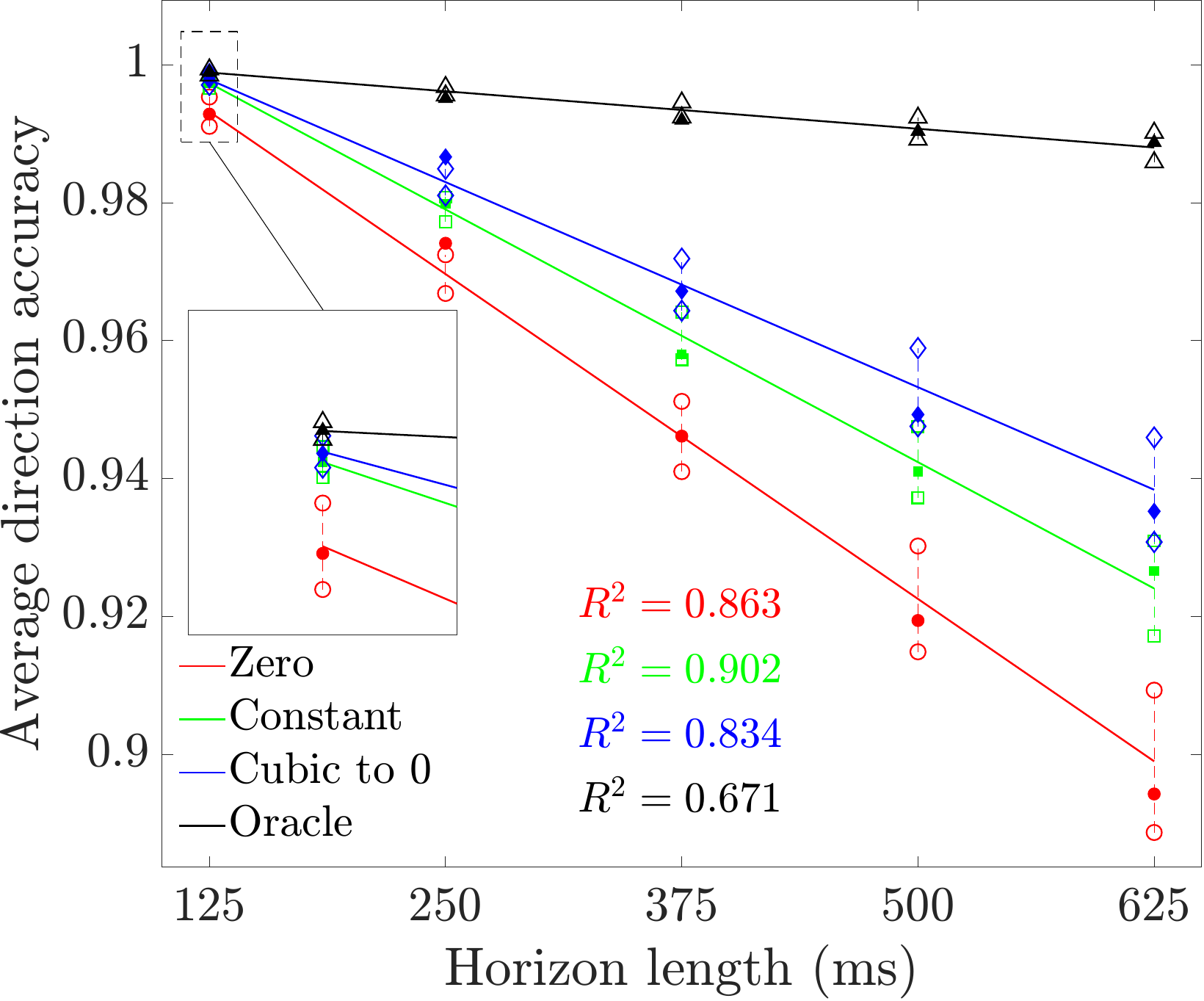}
        \caption{}
        \label{fig:ADA_fit}
    \end{subfigure}\hspace{0.2cm}
    \begin{subfigure}{0.48\textwidth}
        \centering
        \includegraphics[width=\linewidth]{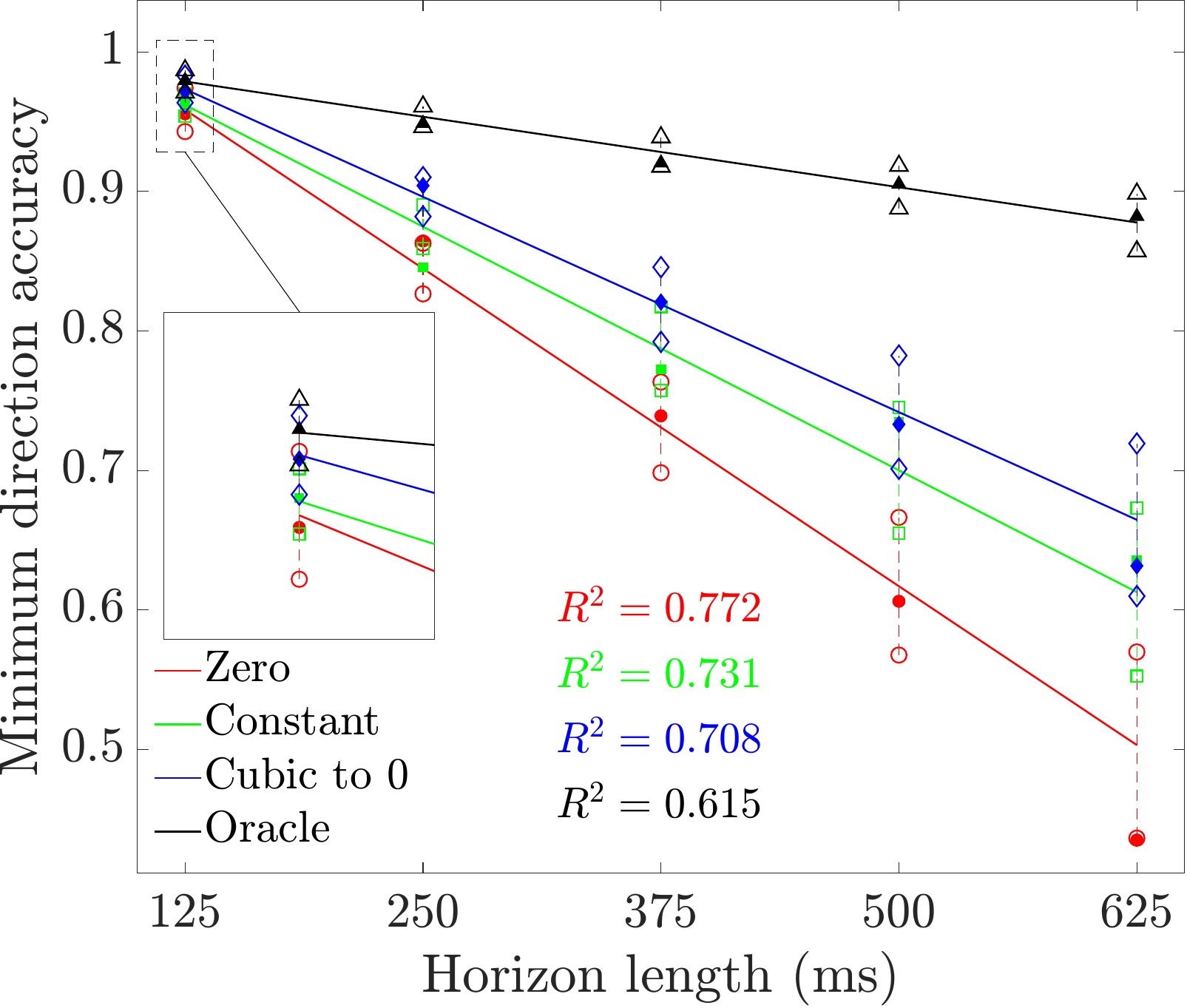}
        \caption{}
        \label{fig:MDA_fit}
    \end{subfigure}
    \caption{The fitted curves for (a) the average direction accuracy ($ADA$), and (b) the minimum direction accuracy ($MDA$). The solid markers (\zerosymb\ for Zero, \constsymb\ for Constant, \cubesymb\ for Cubic to 0, and \oraclesymb\ for Oracle) show the mean of the data points, while the hollow markers represent the 95\% CIs.}
\end{figure}

The outcome of F-tests for $MDA$ did not support the existence of a quadratic trend ($p>0.128$). Among the fitted linear models, the lowest $R^2$ belonged to Oracle ($R^2=0.618$) while the largest $R^2$ was achieved by the Zero profile's fit ($R^2=0.772$) (Fig.~\ref{fig:MDA_fit}). ANOVA found that the acceleration profile had a significant effect on $MDA$ at all horizons ($p<0.030$), but no evidence of a significant difference between Zero, Const, and Cubic pairs was observed in the post-hoc tests (Table~\ref{tab:ADA_MDA_pairs}). For Oracle, the results did not support a difference with any of the other profiles at $T=\SI{125}{\milli\second}$ ($p>0.060$) and with Cubic at  $T=\SI{250}{\milli\second}$ ($p=0.128$); however, the other comparisons were statistically significant ($p<0.025$, $d>1.54$).

\begin{table}[b!]
\renewcommand{\arraystretch}{1}
\fontsize{7.6pt}{9pt}\selectfont
\centering
\caption{The results of post-hoc tests for $AE$ and $ME$ between pairs of acceleration profiles. The effect sizes, represented by Cohen's $d$, are only reported when there is a significant difference.}
\label{tab:AE_ME_pairs}
\begin{tabular}{@{}ccccccclccccc@{}}
\toprule
                             & \multicolumn{1}{l}{} & \multicolumn{5}{c}{Average error ($AE$)}       &  & \multicolumn{5}{c}{Maximum error ($ME$)}       \\ \cmidrule(lr){3-7} \cmidrule(l){9-13} 
\multirow{2}{*}{Pair}        &                      & \multicolumn{5}{c}{$T~(\SI{}{\milli\second})$} &  & \multicolumn{5}{c}{$T~(\SI{}{\milli\second})$} \\ \cmidrule(lr){3-7} \cmidrule(l){9-13} 
                             &                      & 125     & 250     & 375     & 500     & 625    &  & 125      & 250     & 375     & 500    & 625    \\ \cmidrule(r){1-7} \cmidrule(l){9-13} 
\multirow{2}{*}{Zero-Const} &
  $p$ &
  \textbf{$<$0.001} &
  \textbf{$<$0.001} &
  \textbf{0.003} &
  0.695 &
  1 &
   &
  1 &
  0.075 &
  \textbf{0.002} &
  \textbf{$<$0.001} &
  \textbf{$<$0.001} \\
                             & $d$                  & 4.80    & 3.23    & 1.94    & ---     & ---    &  & ---      & ---     & 2.17    & 2.91   & 3.35   \\ \midrule
\multirow{2}{*}{Zero-Cubic} &
  $p$ &
  \textbf{$<$0.001} &
  \textbf{$<$0.001} &
  \textbf{$<$0.001} &
  \textbf{0.004} &
  0.447 &
   &
  \textbf{0.006} &
  1 &
  0.167 &
  \textbf{0.003} &
  \textbf{$<$0.001} \\
                             & $d$                  & 5.18    & 3.98    & 2.91    & 1.85    & ---    &  & 1.742    & ---     & ---     & 1.99   & 2.56   \\ \midrule
\multirow{2}{*}{Const-Cubic} & $p$                  & 1       & 0.303   & 0.208   & 0.227   & 0.213  &  & 0.337    & 0.100   & 0.068   & 0.057  & 0.053  \\
                             & $d$                  & ---     & ---     & ---     & ---     & ---    &  & ---      & ---     & ---     & ---    & ---    \\ \bottomrule
\end{tabular}
\end{table}

\begin{table}[t]
\renewcommand{\arraystretch}{1}
\fontsize{7.6pt}{9pt}\selectfont
\centering
\caption{The results of post-hoc tests for $ADA$ and $MDA$ between pairs of acceleration profiles. The effect size, represented by Cohen's $d$, is reported only when there is a significant difference.}
\label{tab:ADA_MDA_pairs}
\begin{tabular}{@{}ccccccclccccc@{}}
\toprule
\multicolumn{1}{l}{} &
  \multicolumn{1}{l}{} &
  \multicolumn{5}{c}{Average direction accuracy ($ADA$)} &
   &
  \multicolumn{5}{l}{Minimum direction accuracy ($MDA$)} \\ \cmidrule(lr){3-7} \cmidrule(l){9-13} 
\multirow{2}{*}{Pairs} &
   &
  \multicolumn{5}{c}{$T~(\SI{}{\milli\second})$} &
   &
  \multicolumn{5}{c}{$T~(\SI{}{\milli\second})$} \\ \cmidrule(lr){3-7} \cmidrule(l){9-13} 
                             &     & 125            & 250   & 375   & 500   & 625   &  & 125   & 250   & 375 & 500   & 625  \\ \cmidrule(r){1-7} \cmidrule(l){9-13} 
\multirow{2}{*}{Zero-Const}  & $p$ & \textbf{0.014} & 1     & 0.536 & 0.145 & 0.051 &  & 1     & 1     & 1   & 0.274 & 0.21 \\
                             & $d$ & 1.74           & ---   & ---   & ---   & ---   &  & ---   & ---   & --- & ---   & ---  \\ \midrule
\multirow{2}{*}{Zero-Cubic} &
  $p$ &
  \textbf{0.008} &
  \textbf{0.034} &
  \textbf{0.039} &
  \textbf{0.023} &
  \textbf{0.012} &
   &
  0.593 &
  0.602 &
  0.336 &
  0.356 &
  0.249 \\
                             & $d$ & 1.91           & 1.44  & 1.39  & 1.50  & 1.66  &  & ---   & ---   & --- & ---   & ---  \\ \midrule
\multirow{2}{*}{Const-Cubic} & $p$ & 1              & 0.409 & 0.805 & 1     & 1     &  & 0.930 & 0.222 & 1   & 1     & 1    \\
                             & $d$ & ---            & ---   & ---   & ---   & ---   &  & ---   & ---   & --- & ---   & ---  \\ \bottomrule
\end{tabular}
\end{table}

\section{Discussion}

\subsection{Average and maximum error ($AE$ and $ME$)}

For all the profiles, the quadratic fitted curves described the relationship between the prediction error and horizon length $T$ with high accuracy ($R^2>0.918$). For Zero, Const, and Cubic, the trend is due to the discrepancy between the true acceleration and the assumed profile, which manifested in a quadratic error after double integration. This trend can arise if, on average, there is a constant difference between the predicted and actual accelerations (\ref{sect:error}).

The Oracle, despite using the actual GRFs, also exhibited deviations from the whole-body marker set, albeit at a much slower rate of increase compared to the other three. This can be attributed to the drift caused by integration of the noisy GRF signal. Interestingly, a quadratic relationship between prediction error and prediction horizon was also observed in a previous IMU-based  treadmill study \citep{leestma2024data}. The scale of the maximum error ($ME$) was larger than the average ($AE$) by an order of magnitude; for $AE$, it ranged from less than \SI{2.5}{\milli\meter} to \SI{3}{\centi\meter} while the $ME$ reached values of more than \SI{80}{\centi\meter}. This suggests that the performance of the prediction can be highly variable, and, in the longer horizons, unreliable. Whether the assumed acceleration profile provides a close approximation of the true CoM acceleration varies from one horizon to the next, even in the same activity, and also depends on the duration of the horizon (Fig.~\ref{fig:accel_profiles}); however, the behavior of the error can be explained in some special scenarios. For example,  when the CoM experiences low accelerations near zero ($u[k] \approx u[1] \approx 0$) (e.g., the Static activities), all three profiles can approximate the true acceleration similarly well; in cases where $u[1]$ is not close to the subsequent samples, such as the horizons starting near the beginning of a contact, the integration errors can grow to the large values seen in $ME$. The Const profile can be particularly susceptible to significant errors in such cases, which may explain its larger maximum errors (Fig.~\ref{fig:ME_fit}) and the rapid growth of its $AE$ with increasing $T$ (Fig.~\ref{fig:AE_fit}). With longer horizons, the CIs became  wider, showing the higher variability and uncertainty of the error. Consider the Const and Cubic profiles, which only use the first sample ($u[1]$) of the acceleration time series; the shorter the horizon, the more likely the other samples will be similar to $u[1]$, but, as the time grows, the true time series can assume more diverse forms, depending on the activity and how the person performs it, thus leading to the higher spread of errors for larger $T$. 

The post-hoc results in each horizon did not suggest a significant difference between the Const-Cubic pair for either $AE$ or $ME$ (Table~\ref{tab:AE_ME_pairs}). However, the $AE$ of Zero was higher compared to Const and Cubic at $T$ levels up to \SI{375}{\milli\second} and \SI{500}{\milli\second}, respectively. This provides evidence that incorporating the information from the GRFs was beneficial and reduced the average prediction error. In contrast, for $ME$, the Zero profile achieved lower maximum errors than Cubic at horizons lengths of \SI{125}{\milli\second},  \SI{500}{\milli\second}, and \SI{625}{\milli\second}, and lower error than Const at \SI{375}{\milli\second} and above, meaning that at longer horizons, assuming a CoM acceleration of 0 was a better approximation; as discussed previously, Const and Cubic rely on the first acceleration sample of the horizon, which can differ significantly from the rest as the number of samples grows, ultimately leading to worse errors than assuming zero acceleration. The Oracle achieved a smaller error than the other three profiles for all $T$, representing a lower bound on the error, as hypothesized.

\subsection{Average and minimum displacement direction accuracy ($ADA$ and $MDA$)}

The lines for both $ADA$ and $MDA$ were worse fits relative to $AE$ and $ME$ (a minimum $R^2$ of 0.615, Fig.~\ref{fig:MDA_fit}), reflecting the large spread of data points among the subjects; this means the profiles were inconsistent in correctly identifying the main component of intended motion, especially at longer horizons. This is also suggested by the average accuracy ($ADA$) being considerably larger than the worst-case accuracy ($MDA$), with the former remaining close to 0.9, but the latter dropping below 0.5 for the Zero profile at \SI{625}{\milli\second}. Similar to the position errors, a trend of wider CIs with increasing horizon length was also observed for $ADA$ and $MDA$, signaling more uncertainty about whether each profile correctly predicts the main axis of displacement as the prediction horizon becomes longer; this can be explained via the same mechanism discussed for $AE$ and $ME$. Comparing the fitted lines of the individual profiles, it appears that Zero has lower accuracy overall, which decreases faster (i.e., larger slope) than Const and Cubic. The results of post-hoc tests (Table~\ref{tab:ADA_MDA_pairs}) support this observation, with the Zero having a significantly larger $ADA$ than Cubic at all levels of $T$. This provides support that using information from the GRFs, even from one sample at the start of the horizon, could improve the prediction. We note the lack of a significant difference in $MDA$ between any of the pairs, which can be due to the large variations between subjects in this metric. Interestingly, the Oracle's lines also had non-zero slopes, implying that the accumulation of GRF integration error alone also resulted in misidentifying the direction of movement. However, given the small slope of the line, and the lower or similar error at all horizons as shown by post-hoc tests, it still acted as an upper bound for the prediction accuracies. 

\subsection{Recommendations for horizon length acceleration profile}

 The acceptable magnitude of prediction error is an open question and depends on the application; in model predictive control (MPC) the prediction is performed at every update of the control loop, and may be more robust against poor assumptions. For other applications, such as predicting balance, the position of the CoM with respect to the center of pressure and base of support requires more accuracy. Given our results, it is clear that the horizon length is the main determinant of prediction error and accuracy. At $T=\SI{125}{\milli\second}$, the errors and accuracies were close to those attained by the Oracle; interestingly, this horizon length is close to the reaction time of human neuromuscular system in response to disturbances, which may explain its significance as a time scale for human intent \citep{horak1986central, beck2023exoskeletons}. At $T=\SI{250}{\milli\second}$, the profiles also achieved relatively low errors ($AE <$\SI{0.57}{\centi\meter}, $ME <$\SI{11.7}{\centi\meter}) and high accuracies ($ADA >0.97$, $MDA > 0.84$) on average, with narrow CIs. However, at longer horizons, the performance degraded significantly, and, importantly, the CIs became wider, indicating less consistency among tasks and subjects. Therefore, we consider $T=\SI{250}{\milli\second}$ to be a threshold for horizon length.

In the horizons of length 125 and 250\,$\SI{}{\milli\second}$, Const and Cubic profiles, which used the GRFs, achieved mostly better or similar results compared to Zero. There was no significant difference between Const and Cubic in any of these cases; however, Const is more susceptible to large errors, while the assumptions of Cubic produces a more conservative prediction. 

\subsection{Limitations}

While intuitive and interpretable, the acceleration profiles in this work used measurements from only the first sample of the horizon. More complex, data-driven methods, such as neural networks \citep{leestma2024data, molinaro2024task} may improve the prediction, but they require data sets with more activities and subjects than we used to achieve reliable estimates; we believe that ongoing efforts in organizing and producing large-scale biomechanics data \citep{werling2023addbiomechanics, keller2023skin} hold promise for generalizing the presented prediction framework. To perform predictions, we used the CoM state obtained from an optical motion capture system and the 3D GRFs acting on the body recorded by force plates. While these assumptions were necessary for an accurate study of human movement and to establish a reference for prediction errors, implementation on wearable robots without access to drift-free state estimates, and when 3D external forces are not fully available or measurable (e.g., holding onto handrails) remains a future challenge.  

\section{Conclusion}

This work investigated prediction of CoM position during non-cyclic tasks by assuming different profiles for the acceleration in the future. Our analysis revealed that, the position prediction errors grow quadratically with longer horizons, and the accuracy of correctly predicting the main direction of movement decreases linearly. Additionally, we examined the effect of using GRFs and found that it can result in better prediction than assuming constant velocity (i.e., zero acceleration). These results can enable and guide using CoM predictions for applications such as inferring human movement intent.

\FloatBarrier

\bibliographystyle{elsarticle-harv}
\bibliography{ref_updated}

\newpage

\appendix

\section{Description of the activities} \label{sect:activities}

\begin{table}[h!]
\centering
\caption{List of activities performed by participants and the provided instructions. The activities marked with * are in the Static group. Shaded rows are similar to those in the Berg Balance Scale test.}
\label{tab:tasklist}
\small
\begin{tabular}{p{3.2cm}p{8cm}}
\toprule
\textbf{Activity} & \textbf{Description} \\ \midrule
\rowcolor{gray!10}
(1) Sit to Stand & Stand up from the chair without using your hands for support. \\  
\rowcolor{gray!10}
(2) Stand to Sit & Sit down on the chair without using your hands for support.\\
\rowcolor{gray!10}
(3) Foot on Chair & Place your right foot on the upper surface of chair in front of it. The left foot replicates this motion but is not placed on the chair. Repeat for four cycles. \\ 
(4) Stride & Take a stride forward, then stop. \\ 
\rowcolor{gray!10}
(5) Look Behind* & Turn to look directly behind, first over your left shoulder, and then your right shoulder.  \\ 
\rowcolor{gray!10}
(6) Feet Together* & Stand with your feet placed as close together as possible. \\ 
\rowcolor{gray!10}
(7) Eyes Closed* & Stand still with your eyes closed. \\ 
\rowcolor{gray!10}
(8) Turn & Turn 180°, pause, then turn back to your original direction. \\ 
\rowcolor{gray!10}
(9) Tandem Feet* & Stand with one foot directly in front of the other, with your feet as close together as possible. \\ 
\rowcolor{gray!10}
(10) One Leg & Stand on one leg, then return to neutral pose. \\ 
(11) Squat & Perform a squat, then return to neutral pose.\\ 
\rowcolor{gray!10}
(12) Reach Forward & Reach forward as much as possible while keeping your arms parallel, then return to neutral pose.  \\ 
\rowcolor{gray!10}
(13) Object Pickup & Pick up an object placed in front of your right foot. \\ 
(14) Shoe Lace & Step down to emulate tying your right shoe laces, then return to neutral pose. \\ \bottomrule
\end{tabular}
\end{table}

\section{Double-integrator dynamics} \label{sect:dynamics}

We denote the CoM position by $p$, velocity by $\Dot{p}$, acceleration by $\Ddot{p}$, and the GRFs by $R$. If the direction of the $Y$ axis is against gravity, the inputs $u$ are given by 
\begin{gather}
    u_X = \frac{R_X}{m}, u_Y = \frac{R_Y-mg}{m}, u_Z = \frac{R_Z}{m}.
\end{gather}
Therefore, the CoM dynamics is modeled as,
\begin{gather}
    \Ddot{p} = u.
\end{gather}
Defining the states as  $x_1 = p_X$, $x_2 = \Dot{p}_X$, $x_3 = p_Y$, $x_4 = \Dot{p}_Y, x_5 = p_Z$, $x_6 = \Dot{p}_Z$ results in the continuous-time state-state representation below,
\begin{gather}
    \Dot{x} = A_c x + B_c u,\\
        A_c = \begin{bmatrix}
0 & 1 & 0 & 0 & 0 & 0 \\
0 & 0 & 0 & 0 & 0 & 0 \\
0 & 0 & 0 & 1 & 0 & 0 \\
0 & 0 & 0 & 0 & 0 & 0 \\
0 & 0 & 0 & 0 & 0 & 1 \\
0 & 0 & 0 & 0 & 0 & 0 
\end{bmatrix}, B_c = 
\begin{bmatrix}
0 & 0 & 0 \\
1 & 0 & 0 \\
0 & 0 & 0 \\
0 & 1 & 0 \\
0 & 0 & 0 \\
0 & 0 & 1
\end{bmatrix}.
\end{gather}
With the zero-order hold assumption and sample time $\Delta T$, the discrete-time matrices are computed from,
\begin{gather}
    A = e^{A_c \Delta T},\\
    B = \int_0^{\Delta T} e^{A_c t}B_c\,dt.
\end{gather}
This results in,
\begin{gather}
    A = \begin{bmatrix}
1 & \Delta T & 0 & 0 & 0 & 0 \\
0 & 1 & 0 & 0 & 0 & 0 \\
0 & 0 & 1 & \Delta T & 0 & 0 \\
0 & 0 & 0 & 1 & 0 & 0 \\
0 & 0 & 0 & 0 & 1 & \Delta T \\
0 & 0 & 0 & 0 & 0 & 1 
\end{bmatrix}, B = 
\begin{bmatrix}
\frac{\Delta T^2}{2} & 0 & 0 \\
\Delta T & 0 & 0 \\
0 & \frac{\Delta T^2}{2} & 0 \\
0 & \Delta T & 0 \\
0 & 0 & \frac{\Delta T^2}{2} \\
0 & 0 & \Delta T
\end{bmatrix}.\label{eq:ss_A_B}
\end{gather}

\section{Analysis of the quadratic prediction error trend} \label{sect:error}

We illustrate how the difference between the assumed acceleration profile and the true acceleration can lead to quadratic trends in the average error ($AE$) and maximum error ($ME$) as functions of horizon length.

We consider the CoM dynamics in one dimension. The state-space matrices are given by Eq.~(\ref{eq:ss_A_B}),
\begin{gather}
 A = \begin{bmatrix}
1 & \Delta T \\
0 & 1 
\end{bmatrix}, B =  
\begin{bmatrix}
\Delta T^2 \\
\Delta T
\end{bmatrix}.
\end{gather}
From Eq.~(\ref{eq:disc_state}), by propagating the states in time, we can obtain,
\begin{gather}
x[k] = A^{k-1} x[1] + \sum_{i=1}^{k-1} A^{k-1 - i} B\, u[i].
\end{gather}
It can be shown that the matrix products simplify to,
\begin{gather}
 A^{k-1} = \begin{bmatrix}
1 & \Delta T^{k-1} \\
0 & 1 
\end{bmatrix}, A^{k-1 - i} B = 
\begin{bmatrix}
(k-i) \Delta T^2\\
\Delta T
\end{bmatrix}. 
\end{gather}
Since the first component of $x[k]$ represents the position, $p[k]$, we get,
\begin{gather}
    p[k] = p[1] + \Delta T^{k-1} \Dot{p}[1] + \sum_{i=1}^{k-1} (k-i)\Delta T^2 u[i] \label{eq:position_conv}
\end{gather}
From Eq.~(\ref{eq:e_k}), the expression for prediction error simplifies to the absolute difference in the 1D case, 
\begin{gather}
    e[k] = \left| \hat{p}[k] - p[k] \right|.
\end{gather}
Substituting from Eq.~(\ref{eq:position_conv}) we obtain,
\begin{gather}
    e[k] = \left| \sum_{i=1}^{k-1} (k-i)\Delta T^2 ({u_h}[i] - u[i]) \right|,\label{eq:error_full}
\end{gather}
where ${u}_h$ is the simplified acceleration profile, $u$ is the true acceleration, and the initial position and velocity terms cancel out since they are given by the whole-body marker set (Fig.~\ref{fig:pred_framework}). 

Consider the data for a single activity with the total number of sample $N$. Let us denote the total number of horizons in the data by $N_h$. As discussed in the main text, the number of samples in each horizon is shown by $N_s$. To proceed with analysis of average error ($AE$) and maximum error ($ME$), we make the three simplifying assumptions below:

\textbf{Assumption 1}: we assume the difference between $\hat{u}$ and $u$ in the horizon is some constant $c_h$. Plugging into Eq.~(\ref{eq:error_full}) we get,
\begin{gather}
    e[k] = \frac{k^2-k}{2} \Delta T^2 \left| c_h \right|.\label{eq:error_simple}   
\end{gather}

\textbf{Assumption 2}: we assume $N$ is much larger than $N_s$. With this assumption, the number of horizons $N_h$ in the data is not a function of horizon length (i.e., $N_s$). 

\textbf{Assumption 3}: we assume $\left| c_h \right|$ does not depend on $N_s$, and does not vary between horizons for a given $N_s$, meaning that it is the same for all horizons and horizon lengths (i.e.,  $\left| c_h \right| = \left| c \right|$).

From Eqns.~(\ref{eq:AE}) and (\ref{eq:error_simple}), the average error ($AE$) is defined as,
    \begin{gather}
        AE = \text{E}_h \text{E}_s e[k] = \frac{1}{N_h}  \frac{1}{N_s} \sum_{h=1}^{N_h} \sum_{k=1}^{N_s} \frac{k^2-k}{2} \Delta T^2 \left| c \right|
    \end{gather}
We note that $\sum_{k=1}^{N_s} k^2 - k \propto N_s^3 - N_s$, which is quickly dominated $N_s^3$ with increasing $N_s$. Therefore, we get,
\begin{gather}
        AE \propto \frac{1}{N_s} \sum_{k=1}^{N_s} k^2 - k \propto N_s^2 \propto T^2. \label{eq:ae_trend}
    \end{gather}
For $ME$, we have,
    \begin{gather}
        ME = \max\limits_{h} \max\limits_{s} e[k] = \max\limits_{h} \max\limits_{s} \frac{k^2-k}{2} \Delta T^2 \left| c \right|.
    \end{gather}
Due to the accumulation of integration error, it is clear that the maximum $e[k]$ occurs at $k = N_s$. Therefore, $ME$ is determined by $N_s^2 - N_s$, which is dominated by $N_s^2$ as $N_s$ increases,
\begin{gather}
    ME \propto N_s^2 \propto T^2.  \label{eq:me_trend}
\end{gather}
Thus, as shown in Eqns.~(\ref{eq:ae_trend}) and (\ref{eq:me_trend}), under the presented simplifying assumptions, both $AE$ and $ME$ change quadratically with horizon length $T$.

\end{document}